\RequirePackage{amsmath}
\documentclass[runningheads]{llncs}
\usepackage{graphicx}
\usepackage{microtype}
\usepackage{ dsfont }
\usepackage{float}
\usepackage{hyperref}

\usepackage{adjustbox} 
\usepackage{booktabs, tabularx}

\begin{document}

\title{Visual Rationalizations in Deep Reinforcement Learning for Atari Games}
%
%
\author{
Laurens Weitkamp \inst{1}\and
Elise van der Pol\inst{2} \and
Zeynep Akata\inst{2}
}
\authorrunning{ Weitkamp et al.}
%
\institute{Informatics Institute, University of Amsterdam, the Netherlands \and
UvA-Bosch Delta Lab, University of Amsterdam, the Netherlands
}
\maketitle              
\begin{abstract}
Due to the capability of deep learning to perform well in high dimensional problems, deep reinforcement learning agents perform well in challenging tasks such as Atari 2600 games. However, clearly explaining why a certain action is taken by the agent can be as important as the decision itself. Deep reinforcement learning models, as other deep learning models, tend to be opaque in their decision-making process. In this work, we propose to make deep reinforcement learning more transparent by visualizing the evidence on which the agent bases its decision. In this work, we emphasize the importance of producing a justification for an observed action, which could be applied to a black-box decision agent.
\keywords{Explainable AI  \and Reinforcement Learning \and Deep Learning.}
\end{abstract}

%
%
%
\section{Introduction}
\label{sec:intro}
Due to strong results on challenging benchmarks over the last few years, enabled by the use of deep neural networks as function approximators ~\cite{heess2017emergence,mnih2015humanlevel,silver2016mastering} deep reinforcement learning has become an increasingly active field of research.
While neural networks allow reinforcement learning methods to scale to complex problems with large state spaces, their decision-making is opaque
and they can fail in non-obvious ways, for example, if the network fails to generalize well and chooses an action based on the wrong feature.
Moreover, recent work \cite{henderson2017deep} has shown that these methods can lack robustness, with large differences in performance when varying hyperparameters, deep learning libraries or even the random seed. 
Gaining insight into the decision-making process of reinforcement learning (RL) agents can provide new intuitions about why and how they fail. Moreover, agents that can justify with visual elements why a prediction is consistent to a user are more likely to be trusted~\cite{teach1981analysis}. Generating such post-hoc explanations, also referred to as rationalizations, does not only increase trust, but also it is a key component for understanding and interacting with them~\cite{biran2014justification}. 
Motivated by explainability as a means to make the black-box neural networks transparent, we propose to visualize the decision process of a reinforcement learning agent by using Grad-CAM \cite{selvaraju2017gradcam}. 

Grad-CAM creates an activation map that shows prominent spaces of activation given an input image and class, typically in an image classification task. The activation map is calculated through a combination of the convolutional neural network weights and the gradient activations created during a forward pass of the input image and class in the neural network.

Applying this method instead to a reinforcement learning agent, it wil be used to construct action-specific activation maps that highlight the regions of the image that are the most important evidences, for the predicted action of the RL agent.
We evaluate these visualizations on three Atari 2600 games using the OpenAI Gym wrapper, created precisely to tackle difficult problems in deep reinforcement learning. The range of games in the wrapper are diverse in difficulty: they have different long-term reward mechanics and a different action space per game. These difficulties are of interest when looking to explain why the agent takes a specific action given a state.

This paper is structured as follows: The next section, \ref{sec:related}, discusses related works in both reinforcement learning and explainable AI. Section \ref{sec:model} presents the visual rationalization model and explains how it is adapted to reinforcement learning tasks. Following after that is section \ref{sec:exp} which provides the setup required for experiments. This section also provides the results for the rationalization model, including where the model fails. The last section, section \ref{sec:conc}, provides an conclusion to the experiments.

\section{Related Work}
\label{sec:related}
In this section, we discuss previous works relevant to reinforcement learning and explainable artificial intelligence.

\subsection{Deep Reinforcement Learning}
In general, there are two main methods in deep reinforcement learning. The first method uses a neural network to approximate the value function that estimates the value of state, action pairs as to infer a policy. One such value function estimation model is called the Deep-Q Network (DQN), which has had garnered much attention to the field of deep reinforcement learning due to impressive results on challenging benchmarks such as Atari 2600 games \cite{mnih2015humanlevel}. Since the release of this model, a range of modifications have been proposed that have improved this model such as the Deep Recurrent-Q model, the Double DQN model and the Rainbow DQN model \cite{hausknecht2015deep,hessel2017rainbow,van2016deep}. 

The second method used in deep reinforcement learning approximates the policy directly, by parameterizing the policy and using the gradient of these parameters to calculate an optimal policy. This method is called the policy gradient method, and a much cited example of such a method is known as the REINFORCE line of algorithms \cite{williams1992simple}. More recent examples of policy gradient methods include Trust Region Policy Optimization and Proximal Policy Optimization \cite{schulman2015trust,schulman2017prox}.

A hybrid that combines value function methods and policy gradient methods is known as the actor-critic method. In this method, the actor is trying to infer a policy using a state, action pair and the critic is assigning a value to the current state of the actor. In this paper we use the Asynchronous Advantage Actor-Critic (A3C) model which has been used to achieve human level performance on a wide range of Atari 2600 games \cite{mnih2016asynchronous}.

\subsection{Explainable AI}
Generating visual or textual explanations of deep network predictions is a research direction that has recently gained much interest \cite{andreas2016NeuralMN,hendricks2016generating,park2018multimodal,zintgraf2017visualizing}. Following the convention described by Park et al. in \cite{park2018multimodal}, we focus on post-hoc explanations, namely rationalizations where a deep network is trained to explain a black box decision maker which is useful in increasing trust for the end user.

Textual rationalizations are explored in Hendricks et al. \cite{hendricks2016generating} which proposes a loss function based on sampling and reinforcement learning that learns to generate sentences that realize a global sentence property, such as class specificity. Andreas et al. \cite{andreas2016NeuralMN} composes collections of jointly-trained neural modules into deep networks for question answering by  decomposing questions into their linguistic substructures, and using these structures to dynamically instantiate modular networks with reusable components.

As for visual rationalizations, Zintgraf et al. \cite{zintgraf2017visualizing} propose to apply prediction difference analysis to a specific input. \cite{park2018multimodal} utilizes a visual attention module that justifies the predictions of deep networks for visual question answering and activity recognition. In \cite{xrl2} the authors propose to use a perturbation method that selectively blurs regions to calculate the impact on an RL agent's policy. Although this method demonstrates important regions for the agent's decision making, the method used in this paper highlights important regions without the need for such a perturbation method.

Grad-CAM~\cite{selvaraju2017gradcam} uses the gradients of any target concept, i.e. predicted action, flowing into the final convolutional layer to produce a coarse localization map highlighting the important regions in the image for predicting the concept. It has been demonstrated on image classification and captioning. In this work, we adapt it to two reinforcement learning tasks to visually rationalize the predicted action. 

\section{Visual Rationalization Model}
\label{sec:model}
In reinforcement learning, an agent interacting with an environment over a series of discrete time steps observes a state\footnote{Here we assume problems where partial observability can be addressed by representing a state as a small number of past observations} $s_t \in \mathcal{S}$, takes an action $a_t \in \mathcal{A}$ and receives a reward $r_t$ and observes the next state $s_{t+1} \in \mathcal{S}$. The agent is tasked with finding a policy $\pi: \mathcal{S} \times \mathcal{A} \rightarrow [0,1]$, a function mapping states and actions to probabilities whose goal is to maximize the discounted sum of rewards:
\begin{equation}
R_t = \sum^\infty_{k=0} \gamma^k r_{t+k+1}
\end{equation}
which is the return with discount factor $\gamma \in [0, 1]$.

\subsection{Asynchronous Advantage Actor Critic Learning}
Gradient based actor-critic methods split the agent in two components: an actor that interacts with the environment using a policy $\pi(a| s; \theta)$, and a critic that assigns values to these actions using the value function $V(s; \theta)$. Both the policy and the value function are directly parameterized by $\theta$. Updating the policy and value function is done through gradient descent

\begin{equation}
\theta_{t + 1} = \theta_{t} + \nabla_{\theta_{t}} \log \pi(a_t | s_t; \theta_t) A_t.
\end{equation}

With $A_t = R_t - V(s_t ; \theta_t)$, an estimation of the advantage function \cite{Williams1992}. In \cite{mnih2016asynchronous}, the policy gradient actor-critic uses a series of asynchronous actors that all send policy-gradient updates to a single model that keeps track of the parameters $\theta$. In our implementation the actor output is a softmax vector of size $|\mathcal{A}|$, the total number of actions the agent can take in the specific environment. Because our visual rationalization model uses the actor output only, the scalar critic output will be ignored for the purposed of this paper. 
However, in future work, exploring the critic's explanations could be of interest. 
To ensure exploration early on an entropy regularization term $H$ is introduced with respect to the policy gradient,

\begin{equation}
\theta_{t + 1} = \theta_{t} + \nabla_{\theta_{t}} \log \pi(a_t | s_t; \theta_t) A_t + \beta \nabla_{\theta_t} H(\pi (s_t ; \theta_t)),
\end{equation}

where $\beta$ is a hyper parameter discounting the entropy regularization. 


\subsection{Visual Rationalization}
Our visual rationalization is based on Grad-CAM \cite{selvaraju2017gradcam}, and constitutes of computing a class-discriminative localization map $L_{GradCAM}^s \in {R}^{u\times v}$ using the gradient of any target class. These gradients are global-average-pooled to obtain the neuron importance weights $a^c_k$ for class $c$, for activation layer $k$ in the CNN\footnote{$k$ is usually chosen to be the last convolutional layer in the CNN.}:

\begin{equation}
\alpha_k^c = \frac{1}{Z} \sum_i \sum_j \frac{\partial y^c }{\partial L_{ij}^k}.
\end{equation}

\begin{figure}[t]
  \centering
  \includegraphics{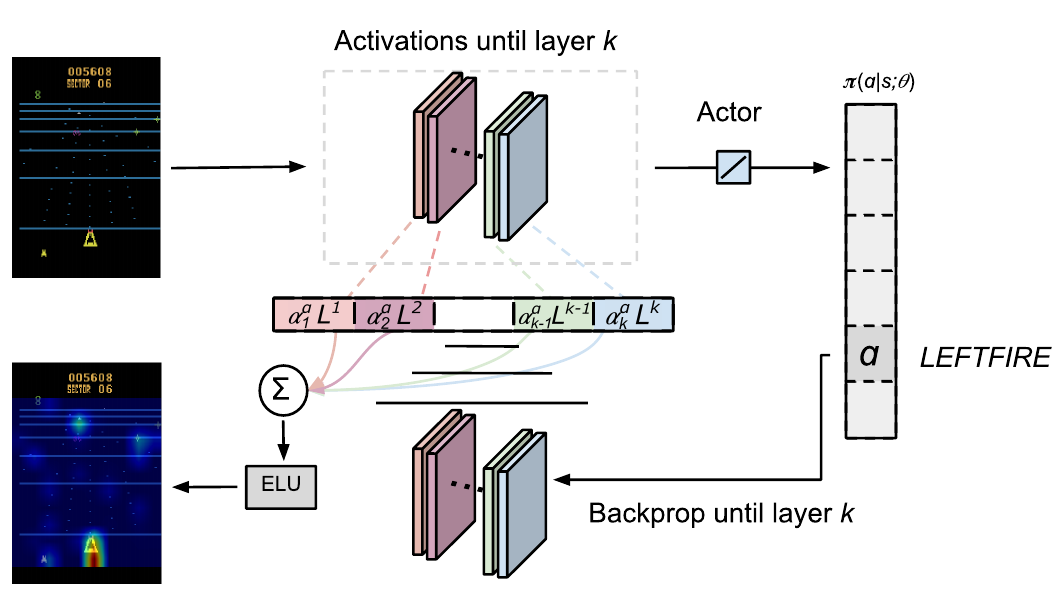}
  \caption{The model takes as input a state, calculates the state-action $\pi (a|s; \theta)$ policy and then produces a gradient-based activation map based on the state, action pair. This activation map can then be overlayed on the original state to indicate evidence that the agent has to take the action. In this Figure, the agent chooses to take the action LEFTFIRE which would make the agent go one step to the left and then shoot up. The activation map is highlighting the agent (bottom), incoming debris (upper-right) and an incoming enemy (upper-mid).}
  \label{fig:A3C_GCAM}
\end{figure}

Adapting this method in particular to the A3C actor output, let $h^a$ be the score for action $a$ before the softmax, $\alpha_k^a$ now represents the importance weight for state $a$ in activation layer $k$: 

\begin{equation}
\alpha_k^a = \frac{1}{Z} \sum_i \sum_j \frac{\partial h^a }{\partial L_{ij}^k},
\end{equation}

with $|h| = |\mathcal{A}|$, the total amount of actions the agent can take. The gradient then gets weighted by the forward-pass activations $L^k$ and passes an ELU activation\footnote{the Exponential Linear Unit has been chosen in favor of the ReLU used in the original Grad-CAM paper due to the dying ReLU effect described in \cite{leaky_relu}.} to produce a weighted class activation map:

\begin{equation}
L^a_{GradCAM} = ELU(\sum_{k=1}^K \alpha^a_k L^k).
\end{equation}
This activation map has values in the range $[0, 1]$ with higher weights corresponding to a stronger response to the input state. This can be applied to the critic output in the same fashion. The resulting activation map can bilinearly extrapolated to the size of the input state and can then be overlayed on top of this state to produce a high-quality heatmap that indicate regions that motivate the agent to take action $a$. A visual representation of this process is depicted in figure \ref{fig:A3C_GCAM}.

\newpage
\section{Experiments}
\label{sec:exp}

In this section, we first provide the details of our experimentation setup. We then show qualitative examples evaluating how our model performs in three of the Atari games. Throughout this section, red bounding boxes and red arrows indicate important regions of a state.

\subsection{Setup}
The Atari 2600 game environment is provided through the Arcade Learning Environment wrapper in the OpenAI Gym framework \cite{DBLP:journals/corr/abs-1207-4708}. The framework has multiple version of each game but for the purpose of this paper the NoFrameskip-v4 environment will be used(OpenAI considers NoFrameskip the canonical Atari environment in gym and v4 is the latest version). Each state is represented as a $210 \times 160 \times 3$ pixel image with a 128-colour palette, and each state is preprocessed to a $84 \times 84 \times 1$ image as input to the network. A side-effect of this preprocessing is that the visual score will be removed from the state in most games, but the agent still gets the reward per state implicitly through the environment.

\begin{figure}[t]
  \centering
  \includegraphics[width = 400pt]{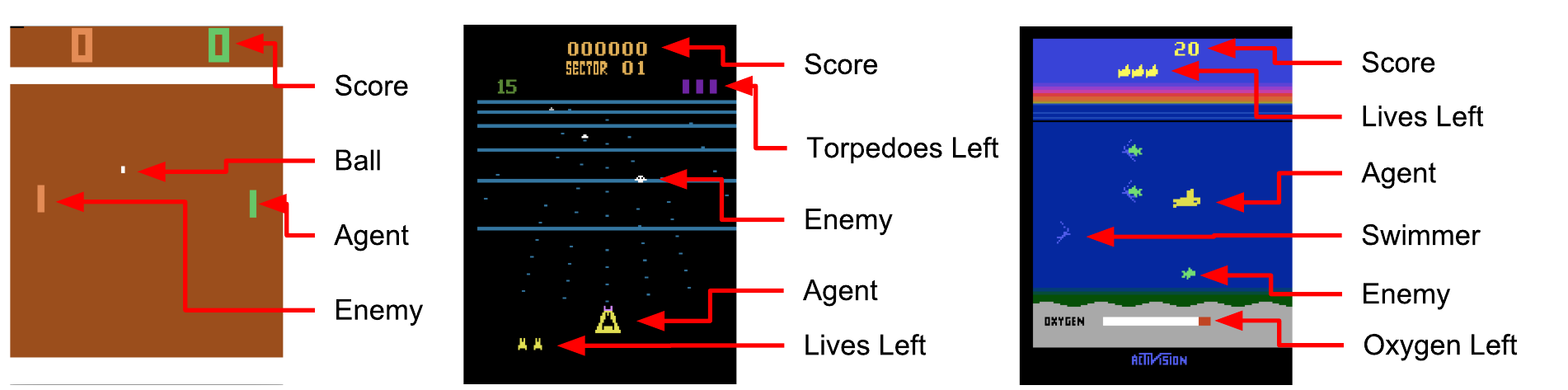}
  \caption{A detailed explanation of the Pong, BeamRider and Seaquest game frames, respectively from Atari 2600 games \cite{DBLP:journals/corr/abs-1207-4708}. The agent is situated in an environment with (multiple) moving enemies, other moving objects and semi-static objects (for example the torpedoes left in BeamRider and the oxygen bar in Seaquest).}
  \label{fig:envs_explained}
\end{figure}

In our experiments, we use three Atari games, namely Pong, BeamRider and Seaquest, all depicted in Figure \ref{fig:envs_explained}. All three games have a different action space (see Table \ref{tab:action_space}), and a different long term-reward system for the agent to learn.

\subsubsection{Pong.} Pong has six actions with three of the six being redundant (FIRE is equal to NOOP, LEFT is equal to LEFTFIRE and RIGHT is equal to RIGHTFIRE). The agent is displayed on the right and the enemy on the left and the first player to score 21 goals wins.

\newpage

\subsubsection{BeamRider.} In BeamRider the agent is displayed at the bottom and the agent has to traverse a series of sectors where each sector contains 15 enemies (remaining enemies is displayed at the top-left) and a boss at the end. The agent has three torpedoes that can be used specifically to kill the sector boss, but these can also be used to destroy debris that appear in later sectors. Learning how to use the torpedoes correctly is not necessary to succeed in the game, but it provides for long-term rewards in the form of bonus points. 

\subsubsection{Seaquest.} In Seaquest the agent is dependent on a limited amount of oxygen, depicted at the bottom of the state. The agent can ascend to the surface which will refill the oxygen bar and it drops off any swimmers that the agent has picked up along the way for a bonus reward. Resurfacing requires learning a long-term reward dependency which is not easily learned \cite{figar}. Surfacing is not just used to refill the oxygen bar but also to drop-off any swimmer that the agent has found underwater which results in additional points. A different way to get a positive reward is to kill sharks.


\begin{table}[t]
\centering
\begin{adjustbox}{max width=\textwidth}
\begin{tabular}{@{}lllllllllllllllllll@{}}
\toprule
                   & NOOP       & FIRE       & UP         & LEFT       & RIGHT      & DOWN       & \begin{tabular}[c]{@{}l@{}}LEFT\\ FIRE\end{tabular} & \begin{tabular}[c]{@{}l@{}}RIGHT\\ FIRE\end{tabular} & \begin{tabular}[c]{@{}l@{}}UP\\ LEFT\end{tabular} & \begin{tabular}[c]{@{}l@{}}UP\\ RIGHT\end{tabular} & \begin{tabular}[c]{@{}l@{}}UP\\ FIRE\end{tabular} & \begin{tabular}[c]{@{}l@{}}DOWN\\ LEFT\end{tabular} & \begin{tabular}[c]{@{}l@{}}DOWN\\ RIGHT\end{tabular} & \begin{tabular}[c]{@{}l@{}}DOWN\\ FIRE\end{tabular} & \begin{tabular}[c]{@{}l@{}}UP\\ LEFT\\ FIRE\end{tabular} & \begin{tabular}[c]{@{}l@{}}UP\\ RIGHT\\ FIRE\end{tabular} & \begin{tabular}[c]{@{}l@{}}DOWN\\ LEFT\\ FIRE\end{tabular} & \begin{tabular}[c]{@{}l@{}}DOWN\\ RIGHT\\ FIRE\end{tabular} \\ \midrule
\textbf{Pong}      & \textbf{x} & \textbf{x} & \textbf{}  & \textbf{x} & \textbf{x} &            & \textbf{x}                                          & \textbf{x}                                           &                                                   &                                                    &                                                   &                                                     &                                                      &                                                     &                                                          &                                                           &                                                            &                                                             \\ \midrule
\textbf{BeamRider} & \textbf{x} & \textbf{x} & \textbf{x} & \textbf{x} & \textbf{x} & \textbf{}  & \textbf{x}                                          & \textbf{x}                                           & \textbf{x}                                        & \textbf{x}                                         &                                                   &                                                     &                                                      &                                                     &                                                          &                                                           &                                                            &                                                             \\ \midrule
\textbf{Seaquest}  & \textbf{x} & \textbf{x} & \textbf{x} & \textbf{x} & \textbf{x} & \textbf{x} & \textbf{x}                                          & \textbf{x}                                           & \textbf{x}                                        & \textbf{x}                                         & \textbf{x}                                        & \textbf{x}                                          & \textbf{x}                                           & \textbf{x}                                          & \textbf{x}                                               & \textbf{x}                                                & \textbf{x}                                                 & \textbf{x}                                                  \\ \bottomrule
\end{tabular}
\end{adjustbox} \\
\caption{Action space of Pong, BeamRider and Seaquest in the Atari 2600 OpenAI wrapper. Each agent from top to bottom has an increasing amount of actions. }
\label{tab:action_space}
\end{table}


\subsection{Learning A Policy}
Training an agent to gain human-like or superhuman-like performance in a complex environment can take millions of input frames. In this section we take the same approach as Greydanus et al. \cite{xrl2}, in which the authors argue that deep RL agents, during training, discard policies in favor for better ones. Seeing how an agent is reacting to different situations at different times of training might make it clear how an agent is trying to maximize long-term rewards. To demonstrate this, two agents have been trained for a different number of frames. The first model which will be called the \textit{Full Agent} has been trained using (at least) 40 million frames. The second agent which will be called the \textit{Half Agent} has been trained using 20 million frames, except for the case of Pong where it has been trained using 500,000 frames, due to the fact that Pong is an easier game to learn. The mean score and variance can be found in Table \ref{table:avg_scores_compared}. For both games a sequence of states were manually sampled, after which both agents have evaluated\footnote{\textit{evaluated} in this case means having forwarded each state that has been manually sampled through the model.} the state to learn spatial-temporal information. States were manually sampled by having a person (one of the authors) play one episode of each game. The states were sampled manually to make sure the samples were not biased towards one agents' policy.

\begin{figure}[h]
  \includegraphics[scale=0.38]{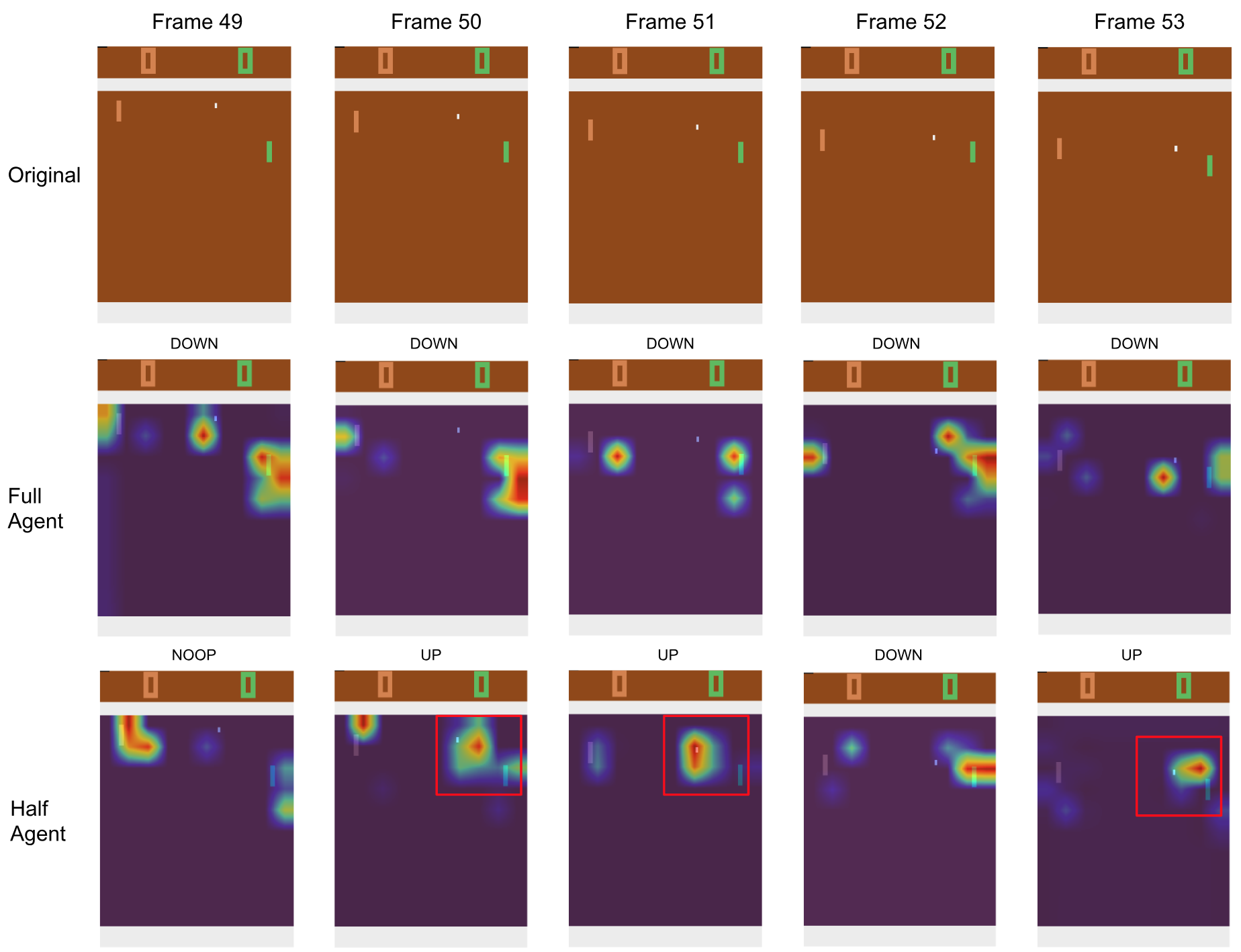}
  \caption{Manually sampled states from the game Pong, combined with the Full Agent and the Half Agent's actions Grad-CAM outputs based on these states. Indicated in the red boxes is the tracking behavior exhibited by the Half Agent. Best viewed in high resolution in color.}
  \label{fig:learning_a_policy_pong}
\end{figure}

\begin{table}[t]
\centering

\begin{adjustbox}{max width=\textwidth}
\begin{tabular}{@{}lrrrrr@{}}
\toprule
                                    & Full Agent Mean & Full Agent Variance & Half Agent Mean & Half Agent Variance \\ \midrule
\textbf{Pong}      & 21.00           & 0.00                & 14.99           & 0.09                &          \\
\textbf{BeamRider} & 4659.04         & 1932.58             & 1597.40         & 1202.00             &          \\
\textbf{Seaquest}  & 1749.00         & 11.44               & N/A             & N/A                 &          \\ \bottomrule
\end{tabular}

\end{adjustbox}
\caption{The mean and variance of both the Full Agent (trained on at least 40 million frames) and the Half Agent (500,000 frames in Pong and 20 million frames in BeamRider) after playing 100 episodes using a greedy strategy. Seaquest's Half Agent is omitted because the Full Agent could not learn how to surface for water.}
\label{table:avg_scores_compared}

\end{table}

\subsubsection{Pong.} For Pong, the Full Agent has learned to shoot the ball in such a way that it scores by hitting the ball only once each round. The initial round might differ, but after that all rounds are the same: the Full Agent shoots the ball up high which makes the ball bounce off the wall all the way down over the opponent’s side, at which point the agent retreats to the lower right corner. This would indicate that the Full Agent is not reacting to the ball most of the time, but is waiting to exploit a working strategy that allows it to win each round. In contrast, he Half Agent is actively tracking the ball at each step and could potentially be losing some rounds because of this. The tracking behavior of the Half Agent is demonstrated in Figure \ref{fig:learning_a_policy_pong} at frames 50, 51 and 53 indicated with a red box. In these frames the Half Agent's attention is focused on the ball and the corresponding action is to go up to match it.

\newpage

\subsubsection{BeamRider.} 
\begin{figure}[h]
    \centering
  \includegraphics[width = 350pt]{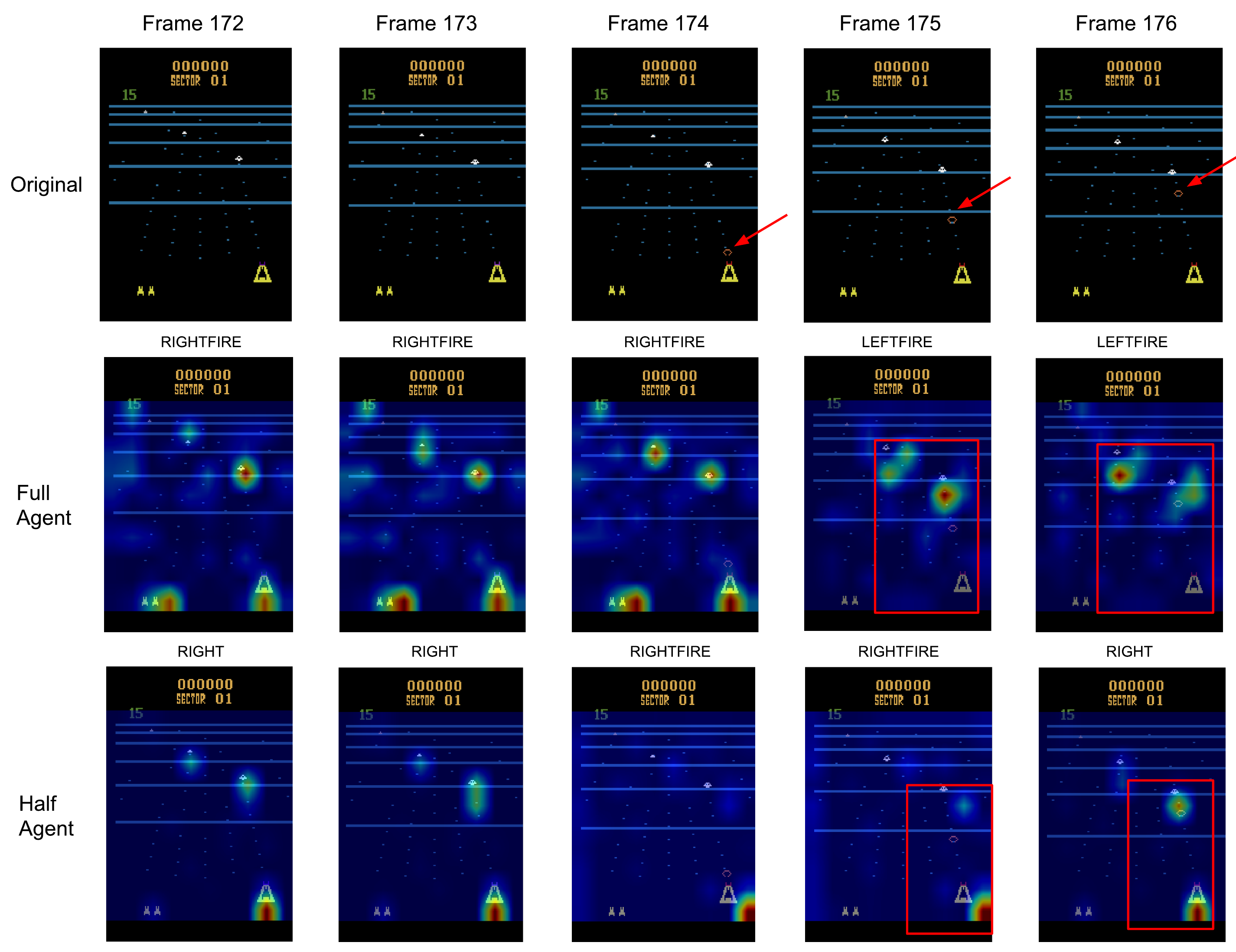}
  \caption{Manually sampled states from the game BeamRider, combined with the Full Agent and the Half Agent's actions Grad-CAM outputs based on these states. The red boxes indicate the difference in focus of the agents and the arrow indicates the shot fired by the agent. Best viewed in high resolution in color.}
  \label{fig:learning_a_policy_beam}
\end{figure}

For BeamRider, both agents have learned to hit enemies but the Full Agent has a higher average return. Looking at figure \ref{fig:learning_a_policy_beam}, both agents have a measure of attention on the two white enemy saucers, but the intensity of attention differs; the Full Agent has high attention on the enemies, in comparison with the Half Agent which has low attention on the enemies. The Half Agent is either going right which is essentially a NOOP in that area or it could be shooting at the incoming enemy. More interesting are the last two frames: 175 and 176. The attention of the Full Agent turns from the directly approaching enemy saucer to the enemy saucer on the left of it, and the agent would try to move into its direction (LEFTFIRE). the Full Agent's attention in frame 176 is placed in a medium degree at the trajectory of its own laser that will hit the enemy saucer in the next frame. This could indicate that the Full Agent knows it will hit the target and is thus moving away from it, to focus on the other remaining enemy.

\begin{figure}[t]
  \includegraphics[scale=0.28]{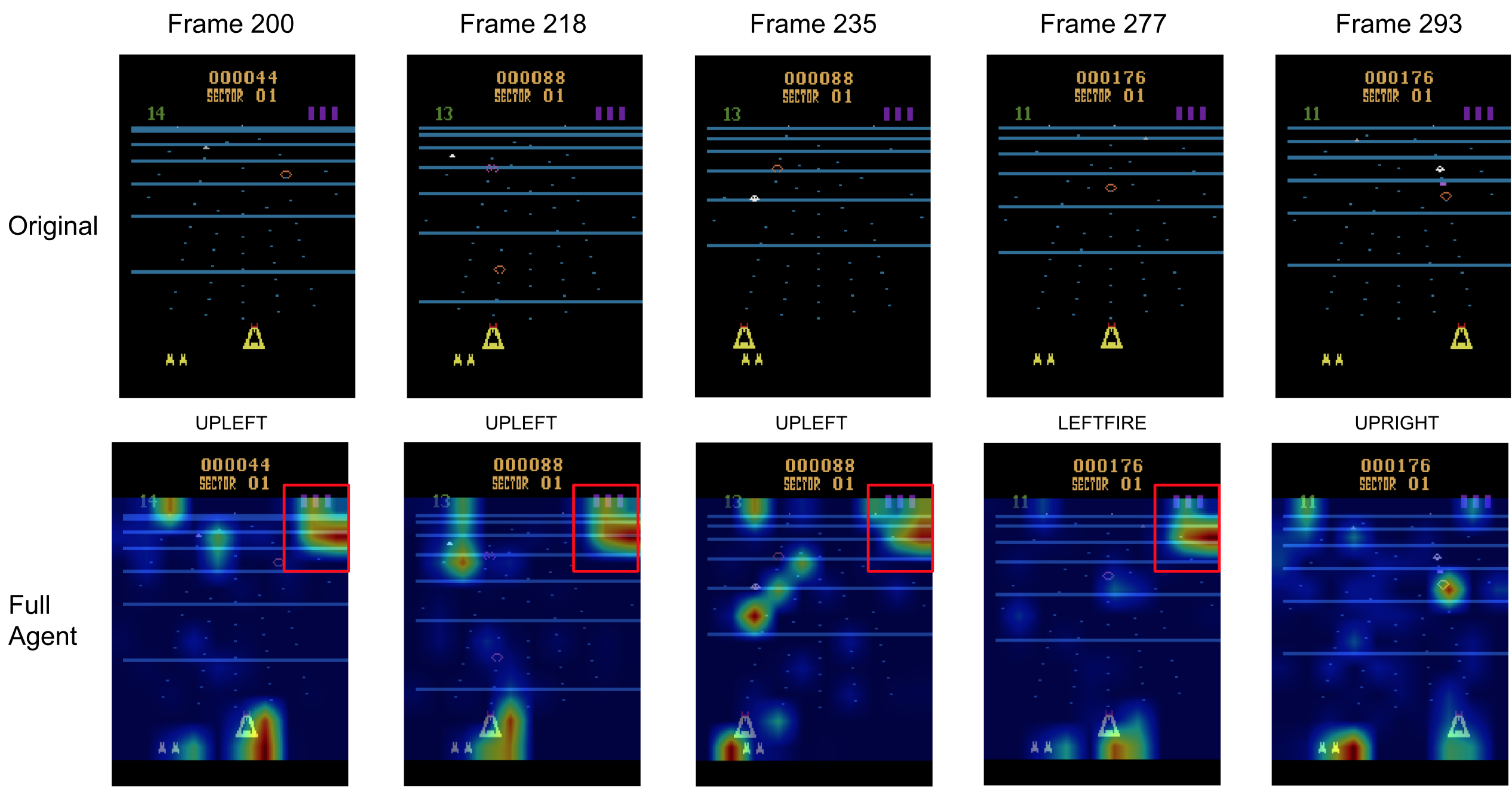}
  \caption{Manually sampled states from the game BeamRider while not firing torpedoes. Combined with the Full Agent's actions Grad-CAM outputs based on these states. In the 300 frames played it has chosen any UP-variant 219 times, LEFTFIRE 67 and other actions 14 times. Best viewed in high resolution in color.}
  \label{fig:learning_a_policy_beam_torpedo}
\end{figure}

From the analysis of both agents another interesting result is discovered: the agents do not learn to \textit{properly} use the torpedoes. At the beginning of each episode/level both agents would fire torpedoes until they are all used up and then continue on as usual. In Figure \ref{fig:learning_a_policy_beam_torpedo} this phenomena is demonstrated through a manually sampled configuration evaluated by the Full Agent only (the results are the same for the Half Agent). The torpedoes have not been used yet, on purpose, and there are enemies coming towards the agent at different time-steps. Looking at the Grad-CAM attention map, it would appear to be highly focused on the remaining three torpedoes in the upper right corner indicated by a red box. This occurs even when the action chosen by the agent is not of the UP-variety which would trigger firing a torpedo.

\newpage
\subsection{Agent Failing}
A different way of looking at how rationalizations aid in understanding the behavior of an agent is by looking at when an agent fails at its task. In the context of BeamRider and Seaquest, this means looking at the last couple of frames before the agent dies. 

\newpage

\subsubsection{BeamRider.} 
\begin{figure}[t]
\centering
    \includegraphics[scale=0.28]{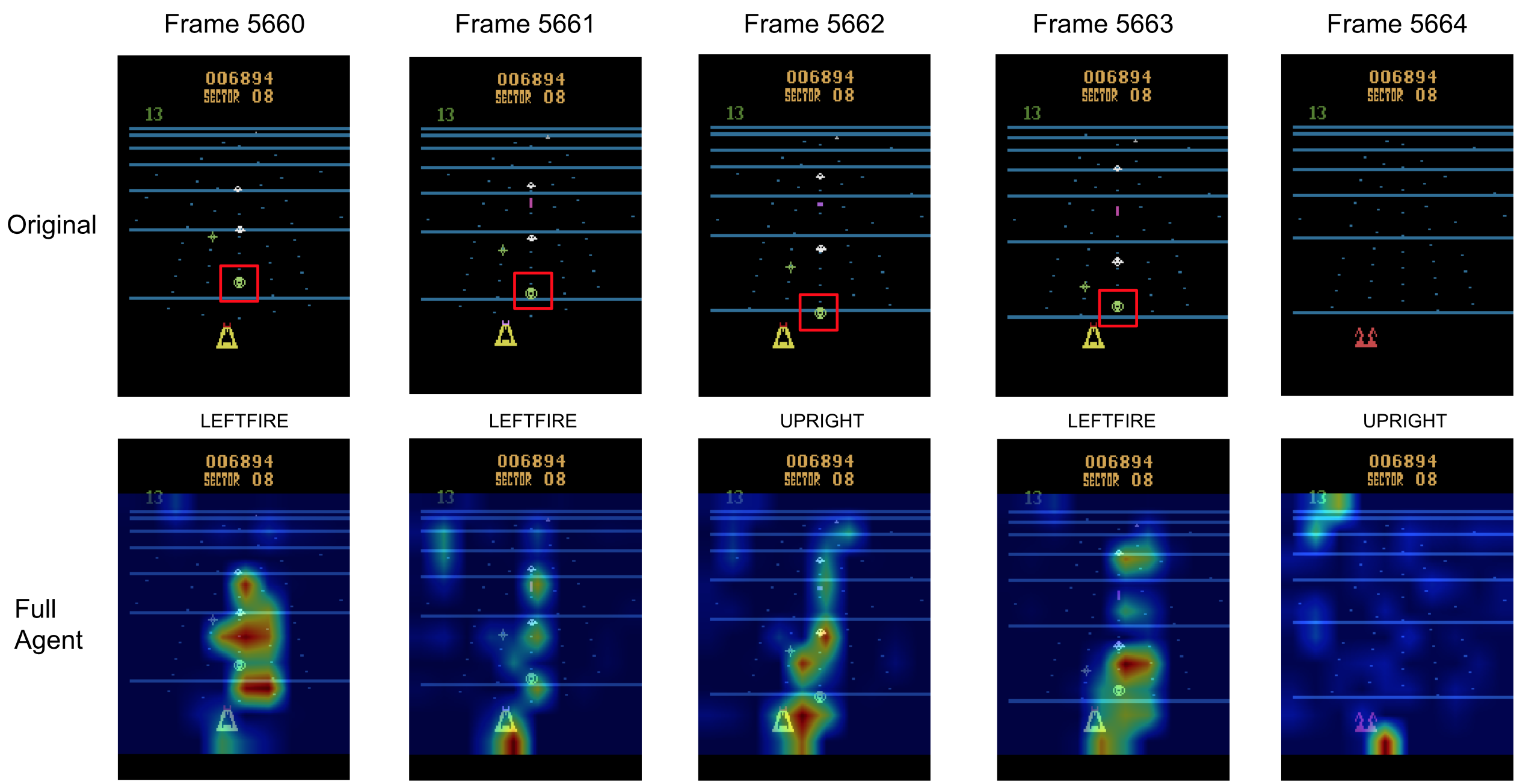}
  \caption{Agent dies because it is hit by a green bouncecraft, highlighted by the rex box. The green bouncecraft is an enemy that only appears in later sectors of the game, but it looks similar to an enemy which is more easily avoidable and which also appears multiple times in each sector. Best viewed in high resolution in color.}
   \label{fig:agent_failures_beamrider.png}
\end{figure}

In the situation depicted in Figure \ref{fig:agent_failures_beamrider.png} the agent is approached by a number of different enemies, one of which only appears after sector 7: the green bounce craft, depicted inside a red box in the first four frames. 
This is an enemy that can only be destroyed by shooting a torpedo at it, and it jumps from beam to beam trying to hit the agent which is what kills the agent eventually in the last frame. 
In all frames the Grad-CAM model is focused at the nearest three enemies, and the agent is shooting using LEFTFIRE in the direction of the green bounce craft.
This could add extra weight to the idea that the agent does not know how to use the torpedoes correctly, but perhaps also that the agent might not be able to distinguish one enemy from another; the piece of green debris to the left of the green bounce craft looks quite similar to it.

\subsubsection{Seaquest.} 
\begin{figure}[t]
\centering
    \includegraphics[scale=0.28]{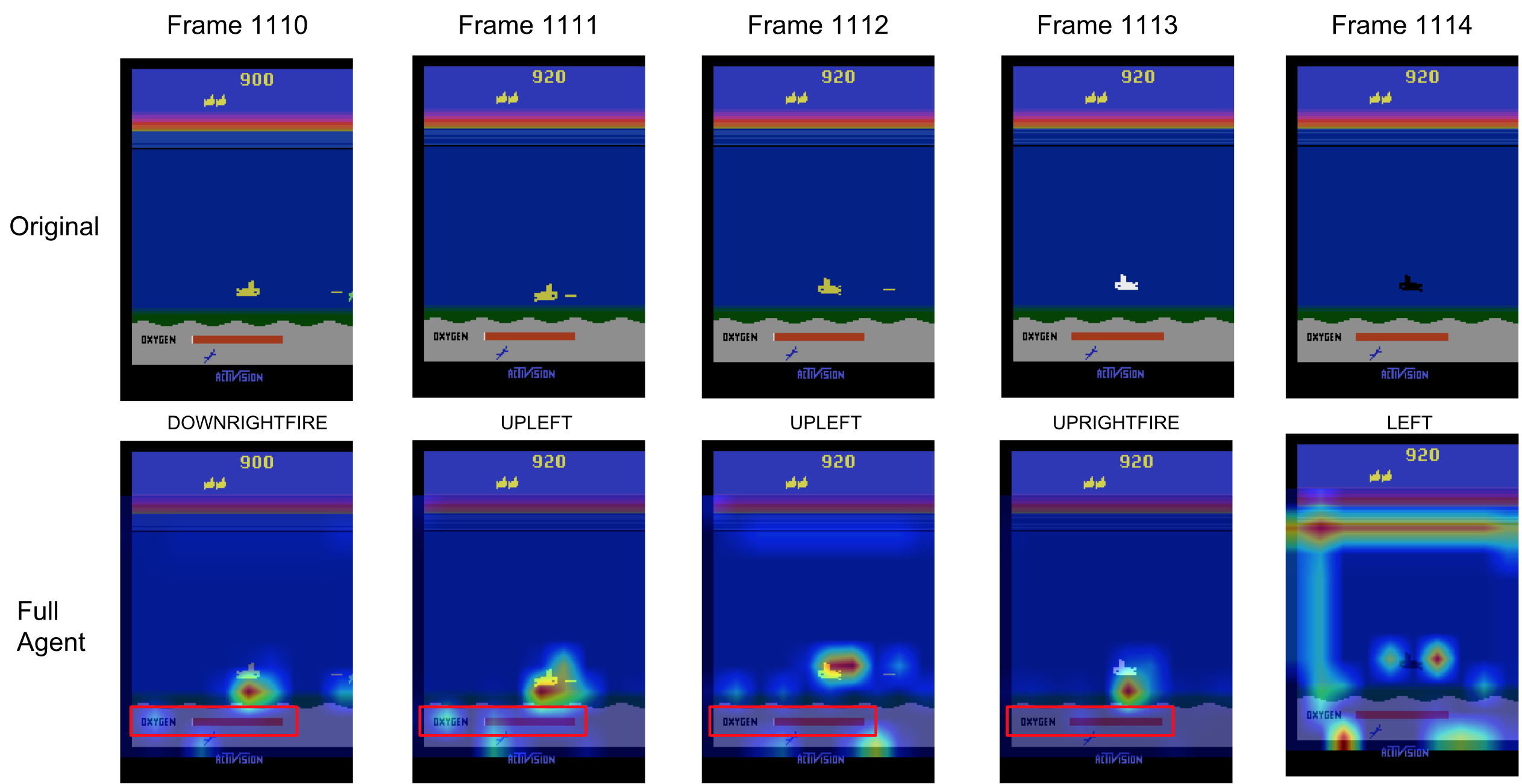}
  \caption{The agent dies due to lack of oxygen depicted in the red box. Looking at the activation map for the Full Agent, it is noticeable that there are no (direct) Grad-CAM activations on the oxygen bar. This could indicate a lack of understanding of the oxygen mechanism that allows the agent to live longer and get a higher score. Best viewed in high resolution in color.}
   \label{fig:agent_failures_seaquest.png}
\end{figure}

The agent playing Seaquest has a different problem: it has not learned the long term strategy of surfacing for oxygen. An example of a death due to this is depicted in Figure \ref{fig:agent_failures_seaquest.png}. The oxygen bar is highlighted by a red box, and it is noticeable that there is no direct or intense activations produced by the rationalization model on the oxygen bar.
This could indicate that the agent has never made a correlation between the oxygen bar depleting and the episode ending.
A multitude of factors could lead the agent to not learn this such as not having enough temporal knowledge or a lack of exploratory actions. A solution to this could be the use of Fine-Grained Action Repetition which selects a random action and performs this action for a decaying number of times \cite{figar}.

\subsection{Failure Cases of Our Model}

Looking at Figure \ref{fig:agent_failures_seaquest.png} a prominent activation is depicted in the form of a vertical bar at the top of the state. This vertical bar might seem a bit too ambiguous and even hard to interpret. 
This type of activation map come in two varieties: activations that highlight only static objects and avoid any non-static objects like agents or enemies and activations that do highlight seemingly at random.

\begin{figure}[t]
\centering
    \includegraphics[scale=0.28]{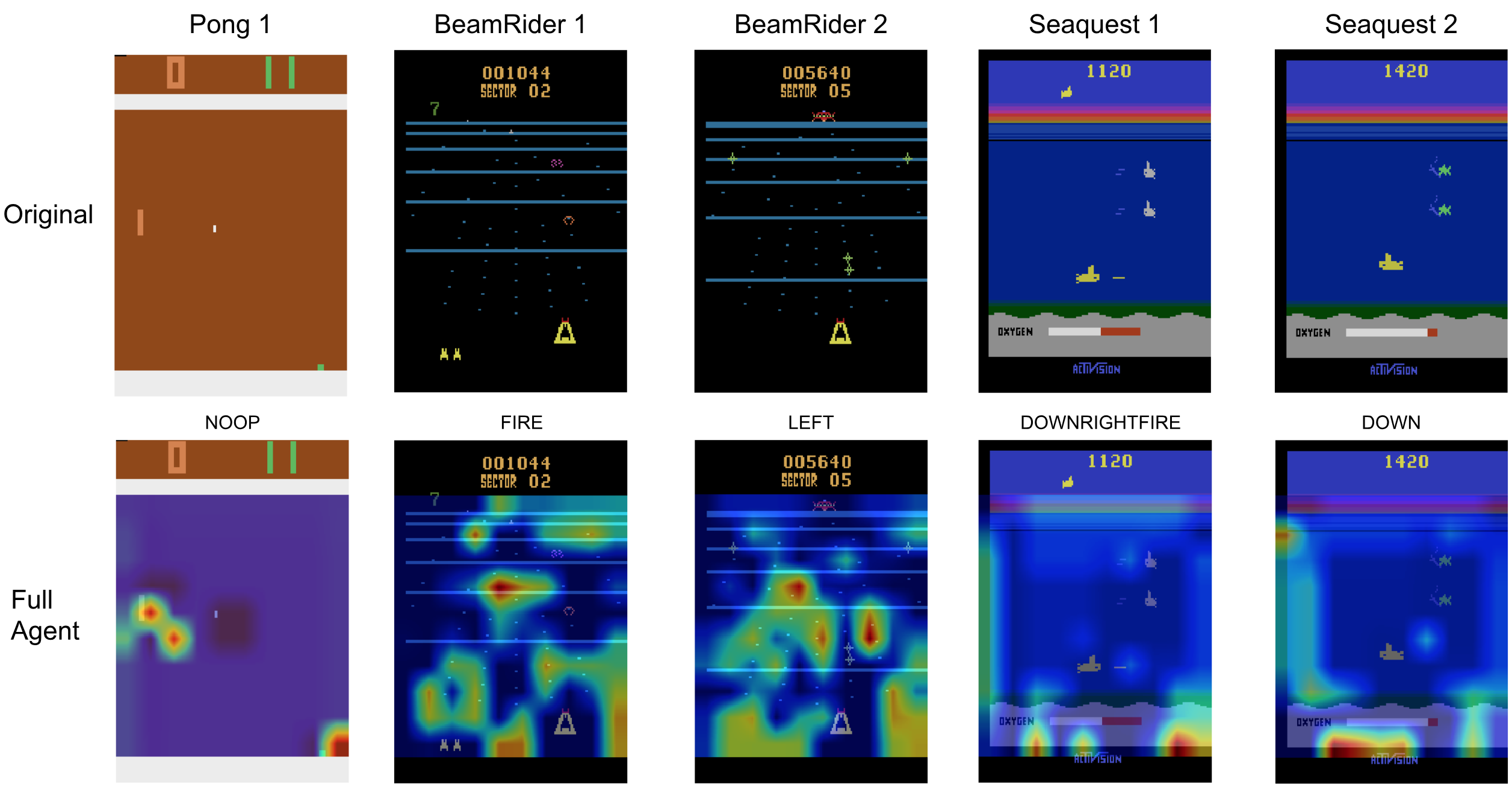}
  \caption{(Seemingly) ambiguous rationalization outputs. The activations depicted in Pong highlight the agent and its enemy. The activations are also noticeable lightly on the whole field except for the bal itself. When looking more closely to the BeamRider activations, it appears that there are activations surrounding important in the game such as the agent, the lives left and incoming enemies. The Seaquest activations, in contrast to the other games, seem more scattered and not focused on either objects or space between objects.
  Best viewed in high resolution in color.}
   \label{fig:interpretability_of_model}
\end{figure}

The activations that highlight everything except for non-static objects are noticeable in Figure \ref{fig:interpretability_of_model} in the case of Pong and BeamRider. For Pong, the activations are not focused on the ball but on everything except for the ball which could still indicate some pattern for the agent. For BeamRider the activations are highlighting areas directly next to the non-static agent and enemies in the state. This could indicate that the agent is calculating the trajectory of enemies or possible safe locations for it to go to.

The activations that are seemingly at random are depicted in Figure \ref{fig:interpretability_of_model} in the last two Seaquest frames. A possible explanation for this could be that the agent is not provided with enough evidence and is indifferent to taking any action, which is reflected in the ambiguous activation map.

\newpage
\section{Conclusion}
\label{sec:conc}
In this work, we have presented a post-hoc explanation framework that visually rationalizes the output of a deep reinforcement learning agent. Once the agent has made the decision of which action to take, the model propagates the gradients that lead to that action back to the image. Hence, it is able to visualize the activation map of the action as a heatmap. Our experiments on three Atari 2600 games indicate that the visualizations successfully attend to the regions such as the agent and the obstacle that lead to the action. We argue that such visual rationalizations, i.e. post-hoc explanations, are important to enable communication between users and the agents. Future work will include a quantitative evaluation in the form of a user study or developing an automatic evaluation metric for these kind of visual explanations.

\newpage
%
%
%
\bibliographystyle{splncs04}
\bibliography{references}

\end{document}